\documentclass[letterpaper, 10 pt, conference]{ieeeconf}  

\IEEEoverridecommandlockouts
\usepackage{graphics} 
\usepackage{amsmath} 
\usepackage{amssymb}  
\usepackage{xcolor} 
\usepackage{xfrac} 
\usepackage{diagbox} 
\usepackage{balance} 
\usepackage{cite} 

\usepackage{multirow} 
\usepackage{array}    

\usepackage{subcaption}  
\usepackage{amsthm}

\usepackage{todonotes}

\usepackage{hyperref} 
\hypersetup{
    colorlinks=true,
    linkcolor=black,
    urlcolor=cyan,
}

\newtheorem{remark}{Remark}

\newcommand{\R}{\mathbb{R}}              
\newcommand{\Spp}{\mathbb{S}_{++}}       

\newcommand{\Xcal}{\mathcal{X}} 
\newcommand{\Ucal}{\mathcal{U}}

\newcommand{\Scal}{\mathcal{S}}

\newcommand{\Qcal}{\mathcal{Q}}

\newcommand{\Dcal}{\mathcal{D}}
\newcommand{\Vcal}{\mathcal{V}}

\newcommand{\T}{\top} 

\newcommand{\nn}{\nonumber}

\renewcommand{\it}[1]{\textit{#1}}

\renewcommand{\rm}[1]{\mathrm{#1}}
\newcommand{\bm}[1]{\mathbf{#1}}    
\newcommand{\bs}[1]{\boldsymbol{#1}} 

\title{\LARGE \bf
Reduced-Order Model Guided Contact-Implicit Model Predictive Control for Humanoid Locomotion
}

\author{Sergio A. Esteban$^{1}$, Vince Kurtz$^{1}$, Adrian B. Ghansah$^{2}$, and Aaron D. Ames$^{1}$
\thanks{
*This research was supported by Technology Innovation Institute (TII).
}
\thanks{
$^{1}$The authors are with the Department of Mechanical and Civil Engineering, California Institute of Technology, Pasadena, CA 91125 USA, {\tt\small\{sesteban, vkurtz, ames\}@caltech.edu}.
}
\thanks{
$^{2}$The authors are with the Department of Control and Dynamical Systems, California Institute of Technology, Pasadena, CA 91125 USA, {\tt\small\{aghansah\}@caltech.edu}.
}
\thanks{Supplemental material: \url{https://rom-cimpc.github.io/}}
}

\begin{document}

\maketitle
\thispagestyle{empty}
\pagestyle{empty}

\begin{abstract}
Humanoid robots have great potential for real-world applications due to their ability to operate in environments built for humans, but their deployment is hindered by the challenge of controlling their underlying high-dimensional nonlinear hybrid dynamics. While reduced-order models like the Hybrid Linear Inverted Pendulum (HLIP) are simple and computationally efficient, they lose whole-body expressiveness. Meanwhile, recent advances in Contact-Implicit Model Predictive Control (CI-MPC) enable robots to plan through multiple hybrid contact modes, but remain vulnerable to local minima and require significant tuning. We propose a control framework that combines the strengths of HLIP and CI-MPC. The reduced-order model generates a nominal gait, while CI-MPC manages the whole-body dynamics and modifies the contact schedule as needed. We demonstrate the effectiveness of this approach in simulation with a novel 24 degree-of-freedom humanoid robot: Achilles. Our proposed framework achieves rough terrain walking, disturbance recovery, robustness under model and state uncertainty, and allows the robot to interact with obstacles in the environment, all while running online in real-time at 50 Hz.
\end{abstract}


\section{Introduction and Related Work}\label{sec:introduction}
Humanoid robots, due to their anthropomorphic structure, are well suited to perform tasks in environments built for humans. 
To realize this potential, humanoid robots must transition from the lab and highly structured settings to real-world environments.  This necessitates whole-body control methods that can replan online in a contact rich fashion---leveraging the environment to facilitate tasks and achieve dynamic stability.  
State-of-the-art whole-body control has advanced significantly in recent years \cite{kuindersma2016optimization, khazoom_humanoid_2022,khazoom2024tailoring}, but high degree-of-freedom (DoF), nonlinear, and hybrid dynamics make whole-body control of humanoid robots a formidable challenge---especially for contact aware real-time controllers.

\begin{figure}
    \centering
    \href{https://rom-cimpc.github.io/}
    {
    \includegraphics[width=\linewidth]{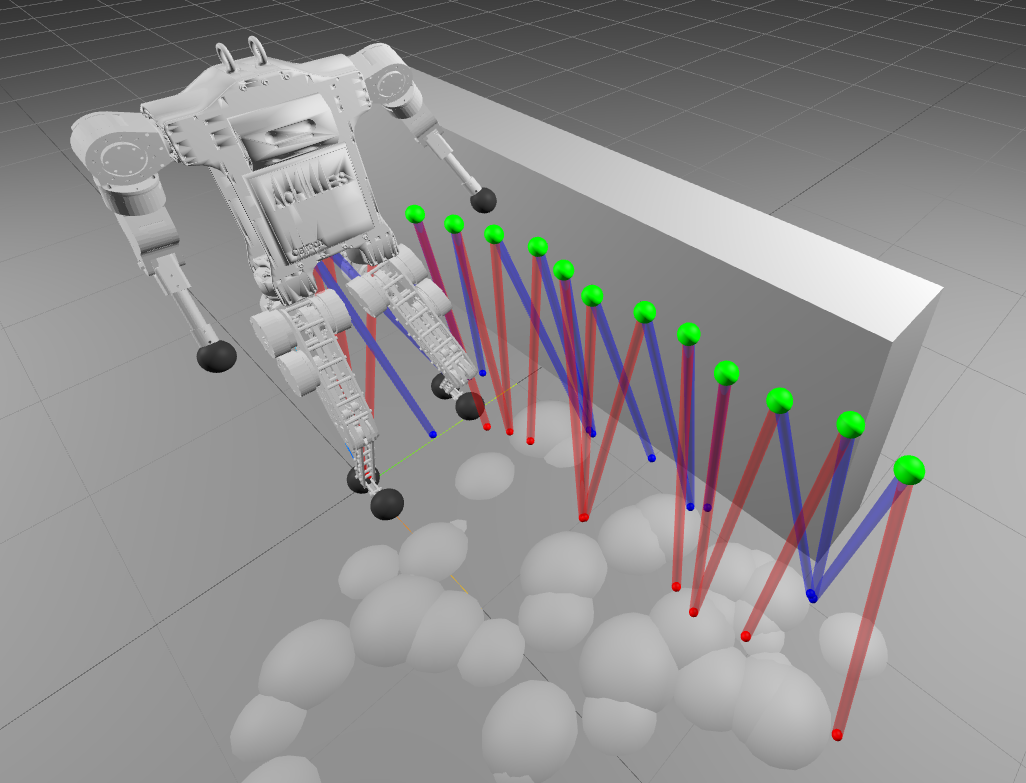}
    }
    \vspace{-14pt} 
    \caption{A simulated Achilles humanoid walks over unmodeled terrain near a wall. The HLIP reduced-order model provides a nominal gait, while CI-MPC adjusts the contact sequence, bracing the arm against the wall.}
    \label{fig:hero}
    \vspace{-18pt} 
\end{figure}
Directly solving the whole-body humanoid optimal control problem is intractable in general. The problem involves not only high-dimensional nonlinear dynamics, but also a power set of potential contact modes. Whole body control becomes even more difficult when the robot is tasked with not only walking over flat ground, but also using its arms and other objects in the environment to move about, as shown in Fig.~\ref{fig:hero}.
To circumvent these optimal control challenges, researchers have developed a rich literature on reduced-order models for locomotion. Point mass models like the Linear Inverted Pendulum (LIP) \cite{kajita20013d}, the Spring Loaded Inverted Pendulum (SLIP) \cite{blickhan1989spring, wensing_generation_2013}, and associated variants \cite{xiong_orbit_2019, dai2024multi,gong2020angular} capture the key characteristics of walking and running. Centroidal dynamics models \cite{orin2013centroidal, dai2014whole} offer a richer characterization while still avoiding the details of each limb's motion. While reduced-order models can achieve robust locomotion and enable fast real-time control, expressiveness at the whole-body level is lost due to abstraction of the full dynamics.

Methods such as \cite{zhao2017multi,hereid20163d,westervelt_hybrid_2003} directly solve the multi-domain trajectory generation problem for the whole body dynamics. To synthesize stable gaits, these methods involve solving difficult nonlinear optimization problems over several possible contact modes. Walking gaits produced with these methods often require an additional foot placement heuristic to mitigate model mismatch. Other methods like \cite{deits2014footstep, ding2020kinodynamic} guide offline trajectory generation with mixed-integer optimization. While the quality of trajectories for these methods are rich, computational requirements limit their usage for settings which require fast replanning.

Model Predictive Control (MPC) offers an alternative approach to the whole-body control problem. MPC addresses the intractability of whole-body control by optimizing over an individual trajectory in receding horizon fashion \cite{wensing_optimization-based_2024, tassa2012synthesis}, rather than considering the whole state-space. Solving the resulting non-convex optimization problem is difficult, and many MPC controllers require a pre-determined contact sequence. Recent work has shown that it is possible to optimize over both the contact sequence and whole-body dynamics in real time \cite{kurtz2023inverse, cleach_fast_2023, aydinoglu2023consensus, kim2023contact}. These Contact-Implicit MPC (CI-MPC) methods leverage specialized solvers to deal with the difficult numerics of contact, and have shown promise for a variety of locomotion and manipulation tasks. Nonetheless, they remain vulnerable to local minima and can require significant parameter tuning, especially for bipedal walking. In practice, direct application of CI-MPC for bipedal locomotion often leads to irregular gaits and foot dragging.

In this paper, we propose a framework that uses the Hybrid Linear Inverted Pendulum (HLIP) \cite{xiong_orbit_2019} to generate reference trajectories for a Contact-Implicit Model Predictive Controller (CI-MPC). This framework inherits the robust foot placement of HLIP while gaining full-body coordination from CI-MPC. HLIP suggests a reasonable contact sequence, guiding CI-MPC towards reasonable  behavior, but the robot is free to deviate from this suggestion or make contact with other objects in the environment.
We demonstrate the effectiveness of this approach in simulation experiments with the 24-DoF Achilles humanoid. HLIP-guided CI-MPC enables versatile and robust locomotion, including walking over rough terrain, improved disturbance rejection, whole-body coordination of legs and arms, walking at different heights, and reaching out to brace against objects in the environment.

The rest of the paper is structured as follows. In Section \ref{sec:preliminary} we cover the background material  for our control framework, which includes a description of the system dynamics, the HLIP model, and CI-MPC. Next, in Section \ref{sec:control_approach} we present our control approach, where we describe how user inputs are mapped to HLIP references, and how these references are passed to our CI-MPC solver to generate stable, expressive behavior. We demonstrate the effectiveness of our approach in Section \ref{sec:results} through several simulation experiments on the 24-DoF humanoid Achilles. In Section \ref{sec:limitations} we discuss limitations of the approach. Finally, in Section \ref{sec:conclusion} we provide some concluding remarks summarizing our findings.



\begin{figure*}[t]
    \vspace{10pt} 
    \centering
    \includegraphics[width=0.95\linewidth]{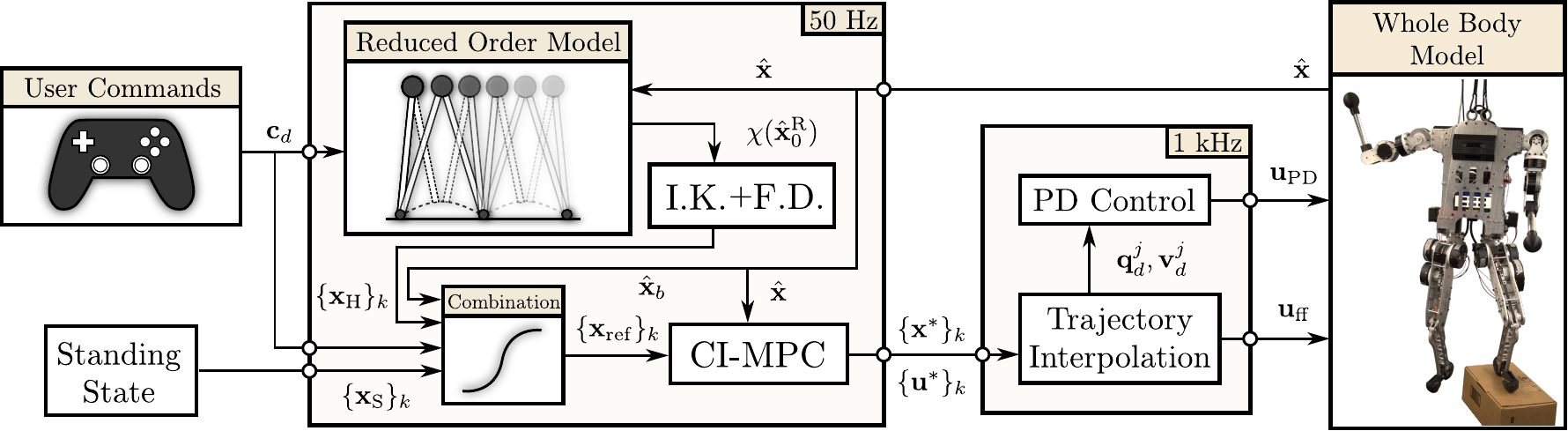}
    \caption{Control architecture that combines the HLIP with CI-MPC. User commands are given to the HLIP. Then, via inverse kinematics and finite difference, we obtain a state trajectory where after combination produces a trajectory for the legs. CI-MPC tracks this trajectory that is interpolated and passed to low-level control of the robot.}
    \label{fig:control_architecture}
    \vspace{-20pt} 
\end{figure*}

\section{Preliminary}\label{sec:preliminary}
\subsection{System Dynamics}
We consider the full 3D dynamics of a humanoid robot. Let $n$ be the number of generalized coordinates of the system. We represent the configuration as:
$
    \bm q = 
    [
    \bm q_b^\T , \bm q_j^\T
    ]^\T
    \in  \Qcal
    \triangleq  SE(3) \times \Qcal_j,
$
and the velocities as:
$
    \bm v = 
    [
    \bm v_b^\T , \bm v_j^\T
    ]^\T
    \in  \Vcal 
    \triangleq \mathfrak{se}(3) \times \Vcal_j,
$
where the subscript $b$ denotes the floating base coordinates and $j$ denotes the joint coordinates that include the arms and legs.
The robot whole body state is then represented as $\bm x~=~[\bm q^\T, \bm v^\T]^{\T} \in \Xcal \triangleq \Qcal \times \Vcal$.

The dynamics of systems that make and break contact with the environment are often modeled by Euler Lagrange equations and holonomic constraints. In this work, we model contact forces via a differentiable function of the robot's state, $\bs \lambda(\bm q)$, as detailed in Section \ref{sec:cimpc}. Thus, our whole body dynamics can be expressed as:
\begin{align} 
    \bm D(\bm q) \dot{\bm v} + \bm{H}(\bm q, \bm v) &= \bm B \bm u + \bm J_c(\bm q)^\T \bs \lambda (\bm q)\label{eq:euler-lagrange} 
\end{align}
where $\bm D: \Qcal \rightarrow \Spp^n$ is the mass-inertia matrix, $\bm H : \Qcal \times \Vcal \rightarrow \R^n$ contains centrifugal, Coriolis, and gravitational terms, $\bm B \in \R^{n \times m}$ is the input selection matrix, and the input vector, $\bm u \in \Ucal \subset \R^m$, includes the joint torques. The contact Jacobian, $\bm J_c: \Qcal \rightarrow \R^{3 n_c\times n}$ maps reaction forces, $\bs \lambda \in \R^{3 n_c}$, to the whole body dynamics, where $n_c$ is the number of contact points with the environment.


\subsection{Hybrid Linear Inverted Pendulum Model}
The HLIP model considers a planar linear inverted pendulum point-mass model with fixed height, $z_0 \in \mathbb{R}$, where the state comprises of the horizontal position, $p \in \mathbb{R}$, and velocity, $v \in \mathbb{R}$, of the point mass relative to the stance foot, $\bm x^{\rm H} = [p, v]^\T$. In this model, the pendulum is assumed to be completely unactuated during the continuous single support phase (SSP). By assuming that the double support period (DSP) is instantaneous, the HLIP dynamics can be expressed as the following single domain hybrid control system,
\begin{equation} \label{eq:hlip_hc}
    \mathcal{HC} 
    =
    \begin{cases} 
        \dot{\bm{x}}_{\rm{ssp}}^{\rm H} = \bm{A}_{\rm{ssp}}\bm{x}_{\rm{ssp}}^{\rm H} & \; \text{if} \quad \bm{x}_{\rm{ssp}}^{\rm H} \in \mathcal{D} \backslash \mathcal{S} \\
        \bm{x}^+_{\rm{ssp}} = \Delta(\bm{x}^-_{\rm{ssp}}) & \; \text{if} \quad  \bm{x}^-_{\rm{ssp}} \in \mathcal{S} 
    \end{cases},
\end{equation}
where $\Dcal$ represents the set of continuous state dynamics and $\Delta: \Scal \rightarrow \Dcal$ is the reset map that maps pre-impact states to post-impact states at the switching surface $\mathcal{S}$. 

The switching of this system corresponds to completing an SSP, i.e., the pendulum falls over but places the next foot at a different location to stabilize $\bm x^\rm{H}$. This introduces the notion of step-to-step dynamics (S2S) which are dynamics considered at the discrete switches. At this level, we can stabilize the HLIP with a foot placement controller to stabilize the S2S dynamics \cite{xiong_3-d_2022}. Achieving walking with this model entails simply tracking the HLIP state. Furthermore, extending the HLIP controller for 3D simply requires an orthogonal composition of two planar HLIPs. We denote the hybrid solution of the 3D HLIP \eqref{eq:hlip_hc}, as:
\begin{equation} \label{eq:HLIP_solution}
    \chi(\bm x_0^{\rm {H}}) = (\Lambda, I, C)
\end{equation}
where $\bm x_0^{\rm {H}}$ is the initial condition, $\Lambda$ is an indexing set for each continuous SSP and $I$ is the set of time intervals for each set of continuous solutions of the SSP, $C$. 


\subsection{Contact-Implicit Model Predictive Control} \label{sec:cimpc}
We aim to achieve full-body locomotion, where the arms and legs coordinate to actively stabilize the robot and interact with the environment. CI-MPC provides a means of generating such motions, including making and breaking contact.


Specifically, given an initial condition $\bm x_0$ and a reference $\bm x_{\rm{ref}}(t)$, CI-MPC solves the trajectory optimization
\begin{subequations} \label{eq:continuous-time-RHC}
\begin{align}
    \underset{\bm u(t)}{\rm {min}} \; \; & \int_{0}^{T} l(\bm x(t), {\bm u(t)}, t) dt + l_f(\bm x(T)), \\
    \rm{s.t.} \; \; 
    & \bm D(\bm q) \dot{\bm v} + \bm{H}(\bm q, \bm v) 
    = \bm B \bm u + \bm J_c(\bm q)^\T \bs \lambda (\bm q, \bm v), \\
    & \bm x(0) = \bm x_0,
\end{align}
\end{subequations}
in receding horizon fashion, where $l(\bm x, \bm u, t) = \| \bm x - \bm x_{\rm{ref}}(t) \|_{\bm Q}^2 + \|\bm u \|_{\bm R}^2$ is a quadratic running cost, $l_f(\bm x) = \| \bm x - \bm x_{\rm{ref}}(T) \|_{\bm V}^2$ is a terminal cost, and $\bm Q, \bm R,\bm V \succeq \bm 0$ are appropriately sized weighting matrices.

While there are various methods to solve this problem \cite{cleach_fast_2023, aydinoglu2023consensus, kim2023contact, jin2024complementarity}, we adopt the Inverse Dynamics Trajectory Optimization (IDTO) approach of \cite{kurtz2023inverse}. IDTO is relatively simple and does not support arbitrary constraints, but its simplicity enables speed and reliability. 

In IDTO, we discretize the trajectory into $N$ steps of size $\Delta t$, and use the generalized positions 
\begin{equation}
    \bm q = [\bm q_0, \bm q_1, \dots, \bm q_N]
\end{equation}
as the only decision variables. 

From generalized positions, we compute velocities and accelerations through forward and backward differences:
\begin{alignat}{2} 
    {\bm v}_k (\bm q) &= \bm N^\dagger(\bm q_k)
    \frac{\bm q_k - \bm q_{k-1}}{\Delta t}, &&\quad\forall k = 1,\dots, N, \label{eq:velocity_q}\\
    \dot {\bm v}_k (\bm q) &= 
    \frac{{\bm v}_{k+1}(\bm q) - {\bm v}_{k}(\bm q)}{\Delta t}, \; \;&&\quad\forall k = 0, \dots, N-1 \label{eq:acceleration_q}
\end{alignat}
where $\bm N^\dagger$ is the kinematic map from generalized position derivatives\footnote{$\bm N(\bm q)^\dagger$ is the left pseudo-inverse of $\bm N(\bm q)$, which maps generalized velocities $\bm v$ to the time derivative of generalized positions $\dot{\bm q}$ \cite{underactuated}.} to generalized velocities, $\bm v = \bm N^\dagger \dot{\bm q}$.
With a compliant contact model \cite{kurtz2023inverse}, the leg and arm contact forces $\bs \lambda$ are also a function of $\bm q$:
\begin{equation} \label{eq:GRF_q}
    \bs \lambda_k (\bm q) = \rm{func} (\bm q_{k+1}, {\bm v}_{k+1}(\bm q)).
\end{equation}
\begin{remark}
    This compliant contact model is used for CI-MPC but not for simulation, where we instead use the physics-based rigid contact model implemented in Drake \cite{drake}.
\end{remark}

Finally, the inverse dynamics \eqref{eq:euler-lagrange} allow us to write generalized forces as a function of $\bm q$ as well:
\begin{multline} \label{eq:inverse_dynamics}
    \bm B \bm u(\bm q) = \bm D(\bm q_{k+1}) \dot {\bm v}_k + 
    \bm H(\bm q_{k+1}, \bm v_{k+1}) \\
    - \bm J_c^\T(\bm q_{k+1}) \bs \lambda_k(\bm q_{k+1}),
\end{multline}
resulting in a compact nonlinear least squares problem where $\bm q$ are the only decision variable. Underactuation (e.g., low-rank $\bm B$) can be handled via constraint or penalty methods \cite{kurtz2023inverse}. 

Importantly, since the contact forces are a smooth function of $\bm q$, we solve for the contact schedules and reaction forces \textit{implicitly}. We use this notion to our advantage to automatically consider different contact modes.





%
\begin{figure*}
    \vspace{10pt} 
    \centering
    \begin{subfigure}[t]{0.49\textwidth}
        \centering
        \includegraphics[width=\textwidth]{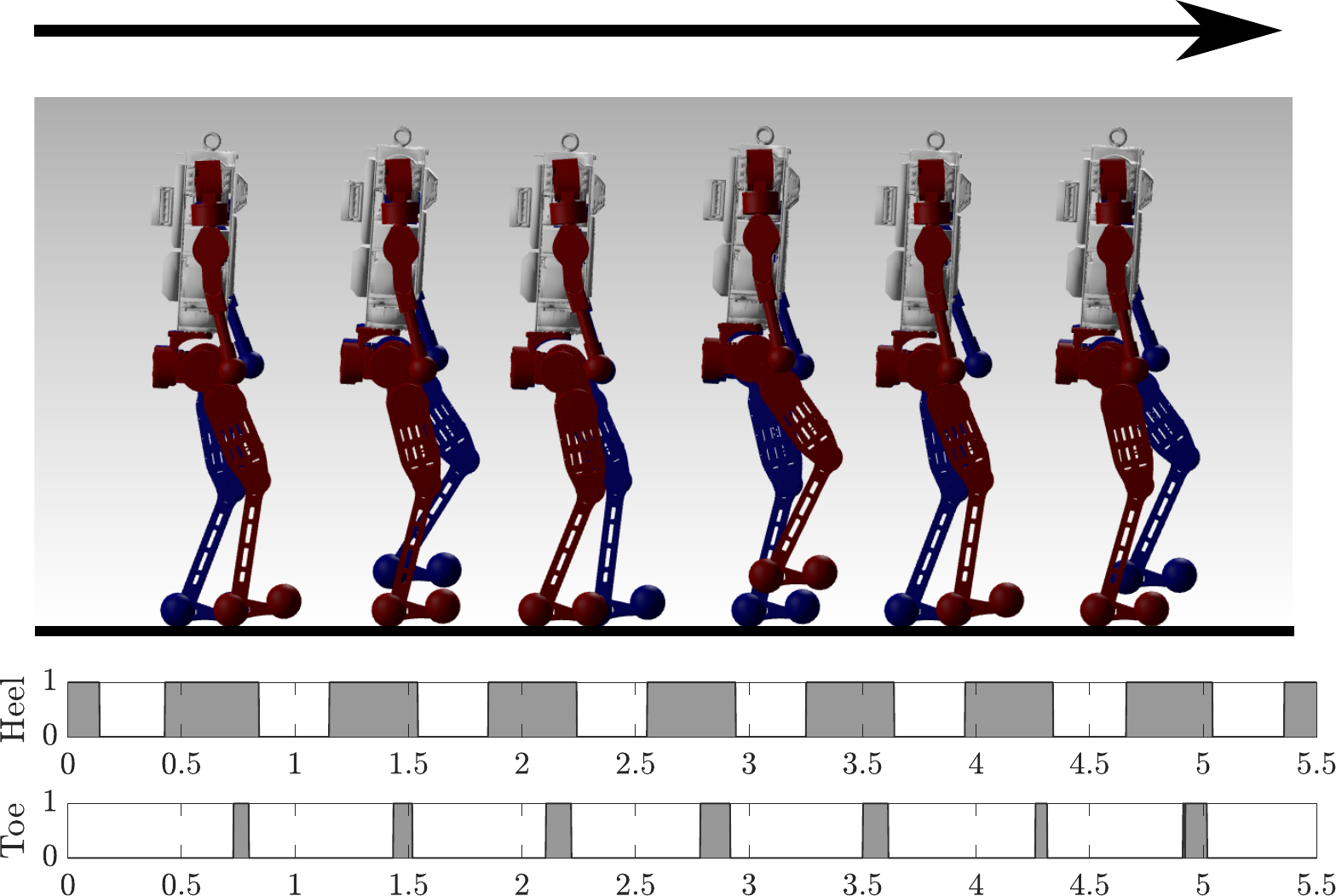}
        \caption{Forward movement}
    \end{subfigure}
    \hfill
    \begin{subfigure}[t]{0.49\textwidth}
        \centering
        \includegraphics[width=\textwidth]{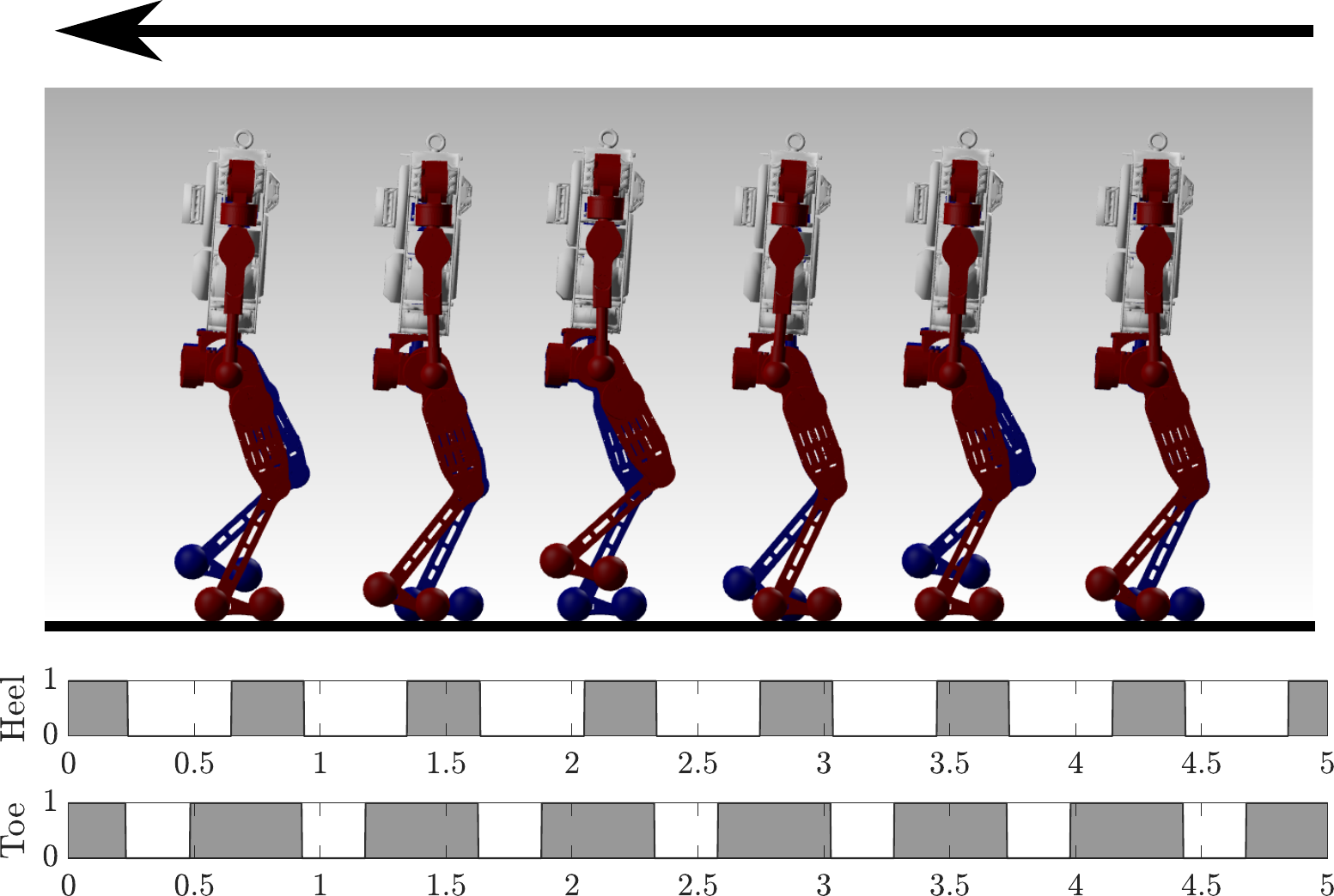}
        \caption{Backward movement}
    \end{subfigure}
    \caption{The proposed framework enables the robot to automatically discover heel-to-toe foot strikes during forward walking and toe-to-heel strikes when moving backward. Shown is an example contact schedule for the right foot when we reach a stable walking gait.}
    \label{fig:banner}
    \vspace{-10pt} 
\end{figure*}

\section{Control Approach}\label{sec:control_approach}
We propose a hierarchical approach to solving the whole-body control problem. The main components of our framework are as follows: (i) receive high-level velocity and height commands from user inputs, (ii) generate a reference trajectory using the HLIP model, and (iii) use CI-MPC to vary contact modes from the reference and generate useful whole-body behavior. This proposed control architecture is shown in Fig. \ref{fig:control_architecture}.
%
\subsection{User Commands}
We define a command vector that includes desired Cartesian velocities, $v_{x,d}, v_{y,d}$, in the robot's body frame, $\{B\}$, desired angular velocity about the $z$-axis, $\omega_{z,d}$, and the desired center of mass (COM) height, $z_{0,d}$ in the world frame, $\{W\}$: 
\begin{equation} \label{eq:vel_commands}
   \bm c_{d} = 
    \begin{bmatrix}
        v_{x,d} & v_{y,d} & \omega_{z,d} & z_{0,d}
    \end{bmatrix} ^\T
     \in \R^4.
\end{equation}
These commands are used to generate a reference trajectory, as described below.



\subsection{Reference Trajectory Synthesis} \label{sec:ref_gen}
We leverage the HLIP model to synthesize useful references for the joint states that are involved with walking. Specifically, we generate HLIP trajectories that track velocity commands $\bm c_d$ given by the user.

Given the robot's current whole-body state, we extract the initial center of mass state, $\bm{x}_0^{\rm{R}}$, which is then used to flow \eqref{eq:hlip_hc} forward in time and obtain $\chi(\bm{x}_0^{\rm{R}})$. The set of continuous solutions, $(\Lambda, I, C)$, provides a stable COM trajectory, but to obtain useful whole body references, we embed the HLIP onto the full body 3D robot kinematics, i.e., we formulate an error tracking problem:
\begin{equation} \label{eq:IK_outputs}
    \bm y =
    \underbrace{
    \begin{bmatrix}
        z^{a}_{\rm{com}} \\
        \bs q^{a}_{\rm{torso}} \\
        \bm p^{a}_{\rm{sw}} \\
        \bs q^{a}_{\rm{sw}}
    \end{bmatrix}
    }_{\bm y^a}
    -
    \underbrace{
    \begin{bmatrix}
        z^{d}_{\rm{com}} \\
        \bs q^{d}_{\rm{torso}} \\
        \bm p^{d}_{\rm{sw}} \\
        \bs q^{d}_{\rm{sw}}
    \end{bmatrix}
    }_{\bm y^d}
\end{equation}
where $(\cdot)^{a}$ and $(\cdot)^{d}$ denote the actual and desired outputs, $z_{\rm{com}}$ is the center of mass height, $\bs  q_{\rm{torso}}$ is the quaternion orientation of the torso, $\bm p_{\rm{sw}}$ is the desired position of the swing foot, and $\bs q_{\rm{sw}}$ is the quaternion orientation of the swing foot. These coordinates are defined with respect to the stance foot frame. The task of driving $\bm y \rightarrow \bm 0$ is achieved via inverse kinematics (IK) which is computed efficiently through the analytical solution. 

As in Section \ref{sec:cimpc}, we discretize the trajectory and solve $N$ inverse kinematics problems:
\begin{equation}\label{eq:inv_kin}
    \bm q^\ell_k = \mathrm{InvKin}(\bm y^d_k), \quad k = 0, 1, \dots, N,
\end{equation}
where $(\cdot)^{\ell}$ denotes the indices of the legs. We obtain leg velocities via finite differences, e.g., 
\begin{equation}
    \bm v^\ell_k = \frac{\bm q^\ell_{k+1} - \bm q^\ell_{k}}{\Delta t}. 
\end{equation}
For the torso coordinates, we propagate an upright configuration tracking the user commands \eqref{eq:vel_commands}. We found that this was more effective than tracking the torso reference from HLIP and inverse kinematics \eqref{eq:inv_kin}. The arm reference was fixed in the initial stationary position at all times. Altogether, this results in an HLIP-based reference for the full system state at each timestep, which we denote $\{ \bm x_{\rm{H}} \}_k$.

%

\subsection{Trajectory Combinations}

To allow gradual transitions between standing and walking, we combine a stable standing reference with the HLIP reference described above. At each MPC replanning iteration, the reference is given by
\begin{equation} \label{ref_combo}
    \{\bm x_{\rm{ref}}\}_k = \alpha \{\bm x_{\rm H}\}_k + (1 - \alpha) \{\bm x_{\rm S}\}_k, 
\end{equation}
where $\{\bm x_{\rm S}\}_k$ is a statically standing trajectory. The scalar $\alpha : \R \rightarrow (0,1)$ weighs the relative importance of the two references, and is defined as
\begin{equation} \label{eq:activation}
\alpha(\phi) = \frac{1}{2} \rm{tanh}(\rho_1(\phi - \rho_2)) + \frac{1}{2},
\end{equation}
where $\rho_1, \rho_2$ are constants and $\phi$ is a function of the commanded velocity, $\bm c_d$, and the current base velocity $\bm v^b$:
\begin{equation}
    \phi = \left\|
    \begin{bmatrix}
        \bm c_d^\T & \bm v_b^\T
    \end{bmatrix} ^\T
    \right\|_{\bm P}^2.
\end{equation}
The norm-defining matrix
\begin{align}
    \bm P &= \mathrm{diag}\Big(
    [v_{x,\mathrm{max}}^{-2}, v_{y,\mathrm{max}}^{-2}, \omega_{z,\mathrm{max}}^{-2}, \\
    &\hspace{1.4cm} z_{0,\mathrm{min}}^{-2}, (v_{x,\mathrm{max}}^b)^{-2}, (v_{y,\mathrm{max}}^b)^{-2}]
    \Big) \nn
\end{align}
specifies a threshold for each component. Here, when one of the velocity components reaches its respective threshold, the weighted norm maps to one. Note that $\phi$ can reach values beyond 1 when multiple velocity components are active simultaneously, but $\alpha(\phi)$ is bounded to $(0, 1)$ by construction.
%
%
\subsection{Contact-Implicit Model Predictive Control} \label{sec:idto}
The blended reference $\{ \bm x_{\rm{ref}}\}_k$ captures leg motion, but does not include useful arm motions and may not be dynamically feasible. CI-MPC takes this reference and generates control actions to track $\{ \bm x_{\rm{ref}}\}_k$ as described in Section~\ref{sec:cimpc}, generating new arm motions and deviating from the nominal contact schedule as needed. 
\subsection{Low-level Control}
CI-MPC outputs an optimal sequence of configurations $\{\bm q^*\}$, however, by construction of the problem, we obtain the optimal velocities via backward difference and the optimal torques via inverse dynamics,
\begin{equation*}
    \{\bm q^*\}_k 
    \quad
    \implies
    \quad 
    \{\bm v^* \}_k, \{\bm u^*\}_k.
\end{equation*}
From here, the solution is interpolated at a higher frequency to obtain the total low-level control input sent to the robot:
\begin{equation}
    \bm u_{\rm{tot}} = \bm u^* + \bm K_p (\bm q_j - \bm q_j^*) + \bm K_d (\bm v_j - \bm v^*_j),
\end{equation}
where $\bm u^*$ is the feed forward term, and $\bm K_{p}$ and $\bm K_{d}$ are PD gains that track the interpolated reference joint positions and velocities $(\bm \{\bm q^*_j\}_k, \{\bm v^*_j\}_k)$.
%


\section{Results}\label{sec:results}

\begin{figure*}
    \vspace{10pt} 
    \centering
    \begin{subfigure}{0.24\linewidth}
        \includegraphics[width=\linewidth]{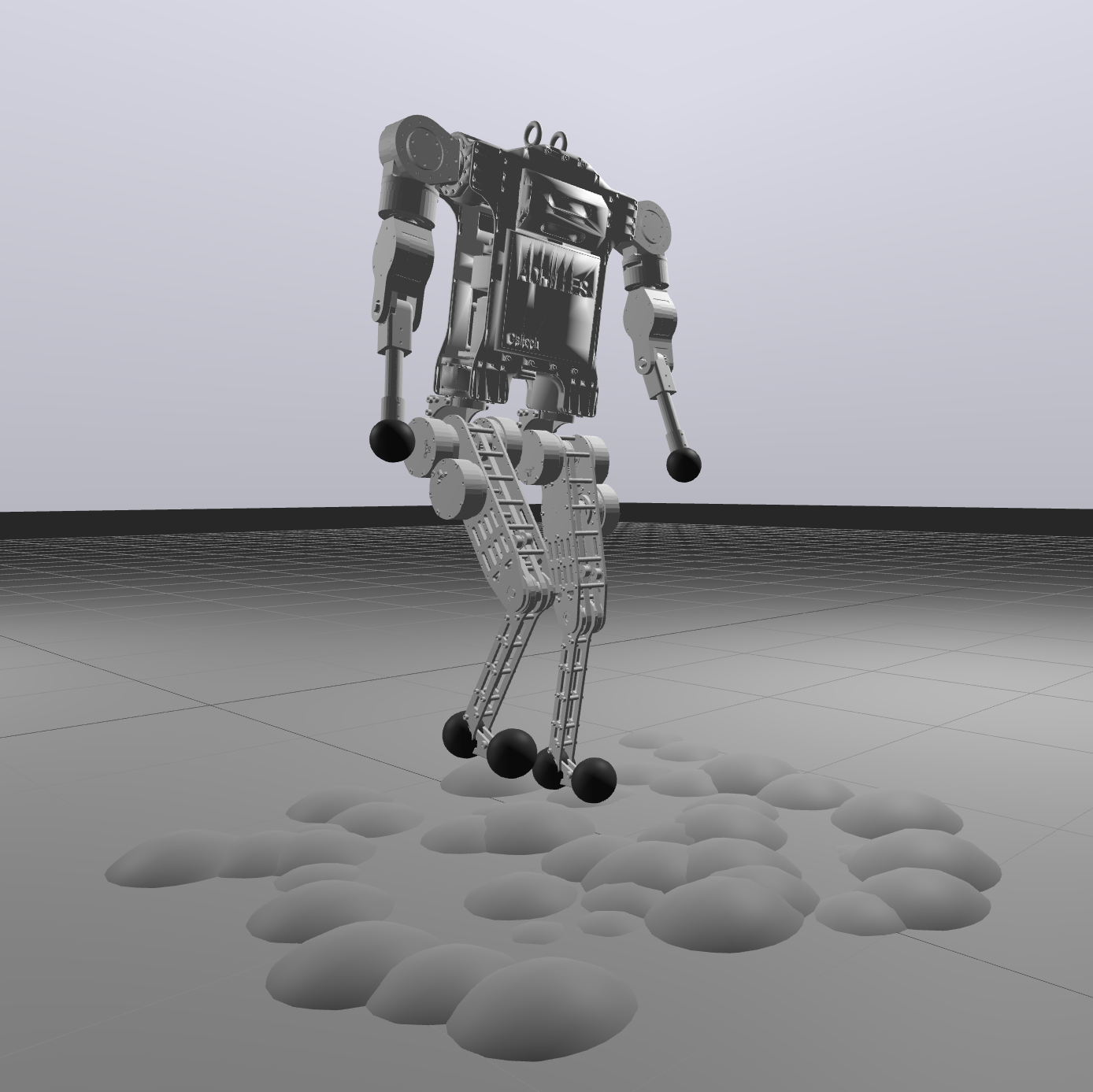}
        \caption{Traversing unmodeled terrain}
        \label{fig:scenarios:terrain}
    \end{subfigure}
    \begin{subfigure}{0.24\linewidth}
        \includegraphics[width=\linewidth]{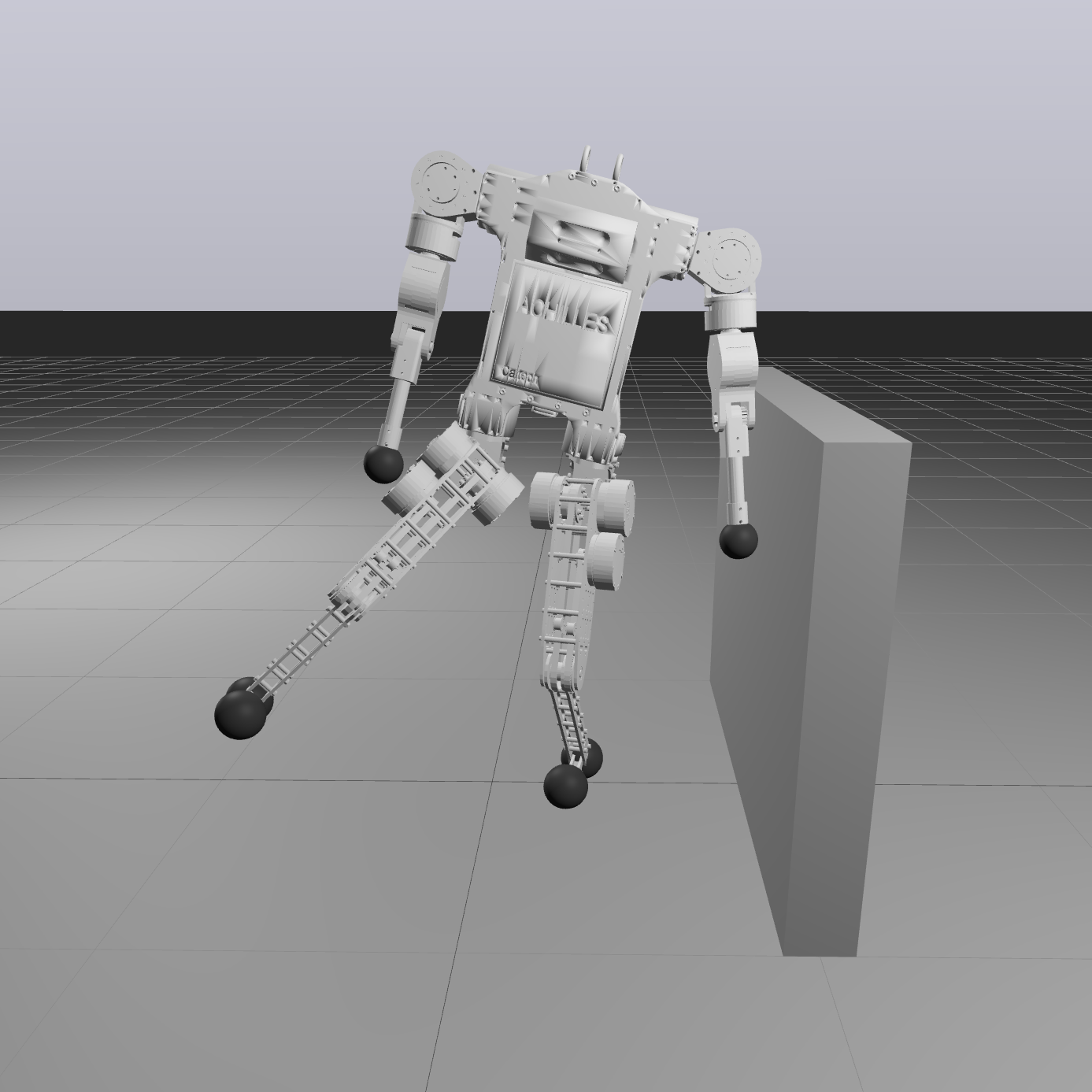}
        \caption{Bracing against a wall}
        \label{fig:scenarios:wall}
    \end{subfigure}
    \begin{subfigure}{0.24\linewidth}
        \includegraphics[width=\linewidth]{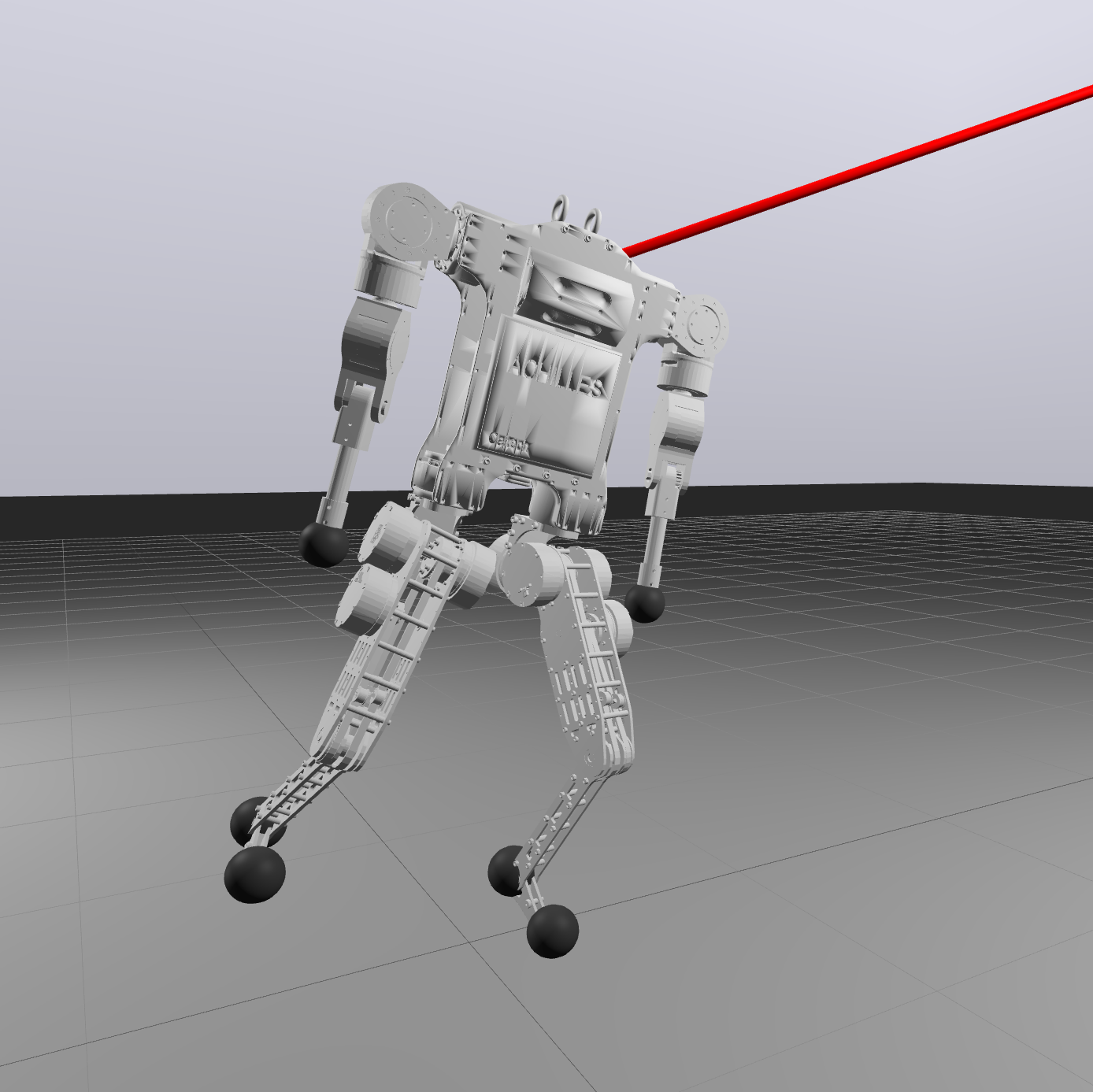}
        \caption{Recovering from disturbances}
        \label{fig:scenarios:wall}
    \end{subfigure}
    \begin{subfigure}{0.24\linewidth}
        \includegraphics[width=0.49\linewidth]{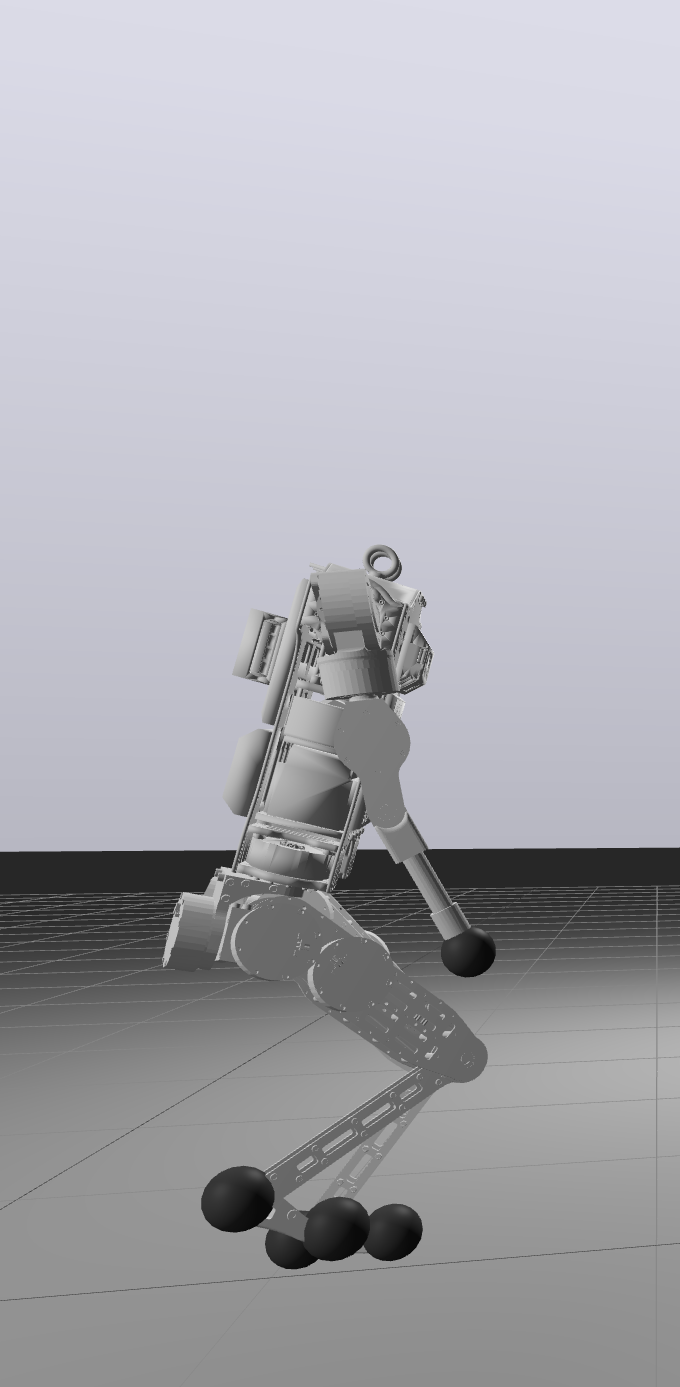}
        \includegraphics[width=0.49\linewidth]{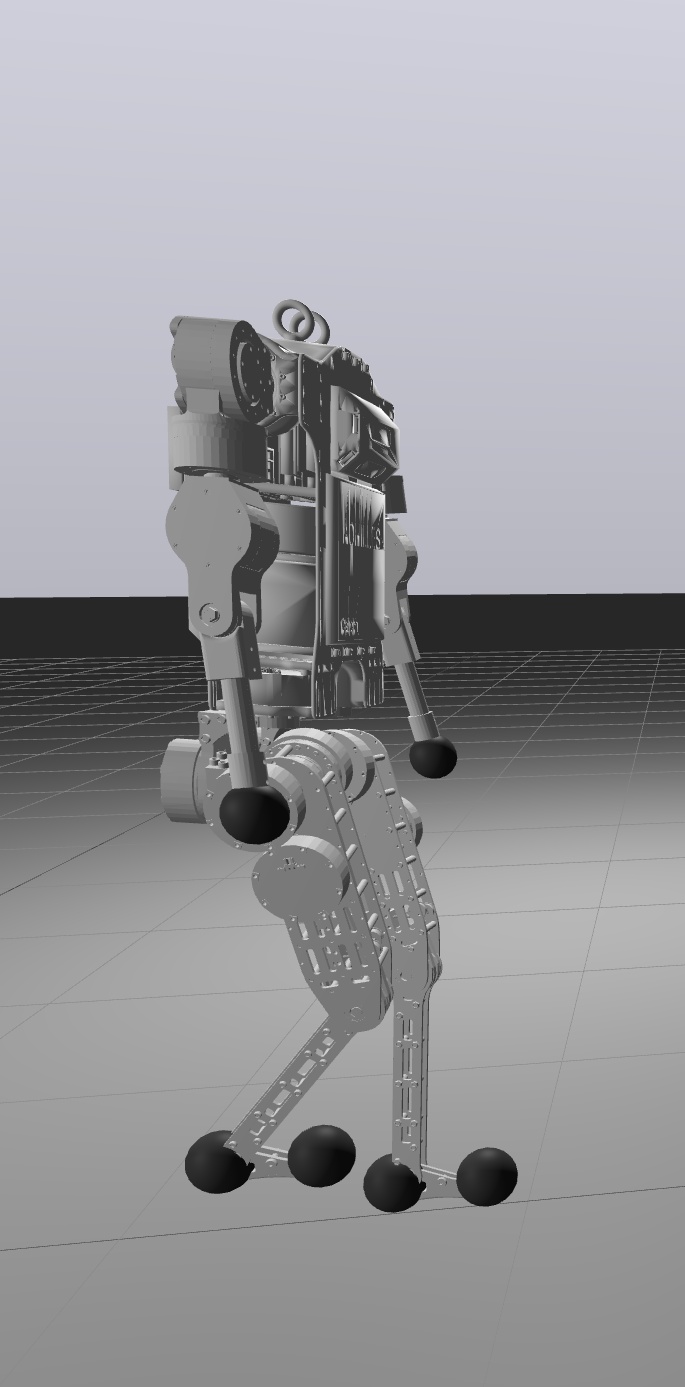}
        \caption{Walking at various heights}
        \label{fig:scenarios:com}
    \end{subfigure}
    \caption{The Achilles humanoid in various simulation test scenarios. HLIP guides the robot toward a reasonable gait, while CI-MPC provides the flexibility to make and break contact on the fly.}
    \label{fig:scenarios}
\end{figure*}

Our proposed HLIP-guided CI-MPC approach enables a diverse array of behaviors, as illustrated in Fig.~\ref{fig:scenarios}. Each of these scenarios are also shown in the supplemental material. 
\subsection{Achilles 3D Humanoid}
We validated our approach on the novel 24-DoF Achilles humanoid platform. Notably, the lack of roll actuation at the feet introduces significant challenges in controlling the underactuated frontal plane dynamics \cite{westervelt2018feedback, buss2014preliminary}. Additionally, by considering the heels, toes, and hands, the robot presents 64 possible hybrid domains. We approximate the robot's contact geometries as spheres as depicted in Fig. \ref{fig:hero}. The proposed framework effectively handles these underactuated dynamics and the large space of contact modes.
\subsection{Simulation Environment}
We test the proposed method in a Drake simulation environment \cite{drake}. Note that while CI-MPC uses a simplified differentiable contact model \eqref{eq:GRF_q}, the simulator uses Drake's state-of-the-art model for rigid contact. We perform all experiments on an Ubuntu 22.04 PC with an AMD Ryzen 9 7950x @ 4.5 GHz, 128 GB RAM. 

\subsection{Parameters}

Key parameters for HLIP and CI-MPC planners are shown in Table~\ref{tab:HLIP_MPC_params}, while parameters for reference blending are listed in Table~\ref{tab:combo_parameters}. These parameters were used in all experiments.

\begin{table}
\centering
\begin{tabular}{c c |c c c c }
    \multicolumn{2}{c}{\textbf{HLIP}} & \multicolumn{4}{c}{\textbf{CI-MPC}} \\
    \hline
    $T_{\rm{ssp}}$ [$s$] & $z_0$ [$m$] & $\Delta t$ [$s$]& $N$ & Freq. [$Hz$] & Iters.\\ 
    \hline
    0.35 & 0.62 & 0.05 & 25 & 50 &  3 \\
    \hline
\end{tabular}
\caption{HLIP and CI-MPC parameters used for experiments.}
\label{tab:HLIP_MPC_params}
\end{table}

\begin{table}
\centering
\setlength{\tabcolsep}{2.5pt} 
\begin{tabular}{c c | c c c c c c}
     \multicolumn{2}{c}{\textbf{Combination}} & \multicolumn{6}{c}{\textbf{Command Thresholds}}  \\
     \hline
        $\rho_1$ & $\rho_2$ &
        $v_x [\frac{m}{s}]$ & $v_y [\frac{m}{s}]$&
        $\omega_z [\frac{\rm{deg}}{s}]$& $z_0 [m]$&
        $v_{x,b} [\frac{m}{s}]$ & $v_{y,b} [\frac{m}{s}]$ \\
    \hline
    5.0 & 0.5 & 0.7 & 0.3 & 35.0 & 0.45 & 0.5 & 0.4 \\
    \hline
\end{tabular}
 
\caption{Parameters for $\alpha(\cdot)$ and maximum commands and base velocities considered in the diagonals of matrix $\bm P$.}
\label{tab:combo_parameters}
\end{table}

\begin{table}
    \centering
    \begin{tabular}{c|c c c}
         \textbf{Joint} & \textbf{Position} & \textbf{Velocity} & \textbf{Torque}  \\
         \hline
         Base orientation & 4.0 & 0.4 & 100.0 \\
         Base position & 1.0 & 0.1 & 100.0 \\
         Hip & 0.4 & 0.03 & 0.001 \\
         Knee & 0.3 & 0.03 & 0.001 \\
         Ankle & 0.8 & 0.08 & 0.0005 \\
         Shoulder & 0.1 & 0.01 & 0.005 \\
         Arm yaw & 0.01 & 0.001 & 0.005 \\
         Elbow & 0.01 & 0.003 & 0.005 \\
    \end{tabular}
    \caption{CI-MPC cost weights}
    \label{tab:mpc_weights}
\end{table}

Table~\ref{tab:mpc_weights} reports values of the diagonal cost weights matrices $\bm Q$ and $\bm R$ for CI-MPC. We use terminal cost weights $\bm V = 10 \bm Q$. Note that our CI-MPC solver, due to its inverse dynamics formulation, requires a penalty on torques on underactuated degrees of freedom like the base, see \cite{kurtz2023inverse} for details.

\subsection{Walking Results}
We begin with flat-ground walking. Applying CI-MPC alone results in a gait with significant foot dragging. Our proposed HLIP guidance results in the more natural gait shown in Fig.~\ref{fig:banner}. A heel-toe contact sequence emerges automatically during forward walking, while backward walking produces a toe-heel sequence. Furthermore, arm swinging emerges at high velocities and low walking heights.

\begin{figure} 
    \centering
    \includegraphics[width=0.8\linewidth]{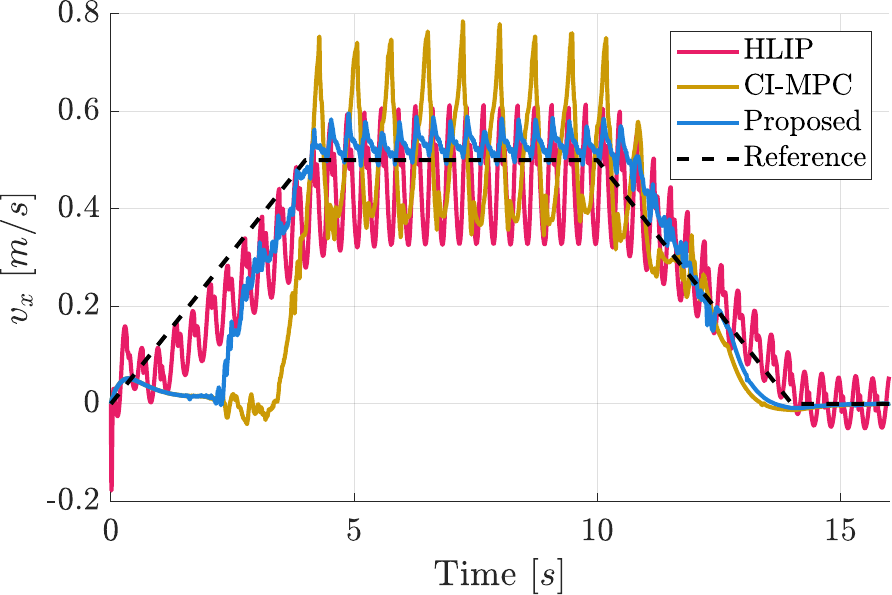}
    \caption{Velocity tracking with HLIP only, CI-MPC only, and our proposed approach that combines the two. CI-MPC only and our proposed approach both take some time before moving forward, as the lowest-cost behavior at small velocity commands is to remain standing in place. Additionally, at low velocities, the the standing configuration reference dominates in \eqref{ref_combo}.}
    \label{fig:velocity_tracking}
    \vspace{-20pt} 
\end{figure}

Velocity tracking performance over flat ground is shown in Fig.~\ref{fig:velocity_tracking}. Our proposed combination of HLIP and CI-MPC provides tighter velocity tracking with less oscillation than either HLIP or CI-MPC alone. Note that HLIP alone continues to move back and forth after stopping, while the reference blending described in Sec.~\ref{sec:control_approach} allows the robot to come to a smooth and complete stop.

\subsection{Disturbances}
\begin{figure}
    \vspace{10pt}
    \centering
    \includegraphics[width=1.0 \linewidth]{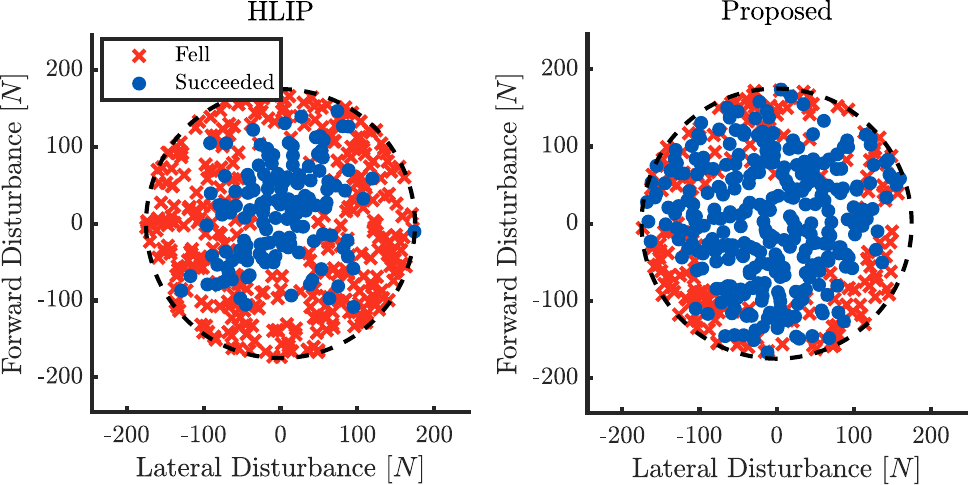}
    \caption{Push recovery comparison between HLIP only (left) and our proposed approach (right). A disturbance force was applied to the base for 0.1 seconds.}
    \label{fig:disturbance_stats}
    \vspace{-10pt} 
\end{figure}

Next, we consider the robustness of our proposed approach to external disturbances. To measure this, we apply randomly generated forces in the horizontal plane to the base of the robot and record whether the robot remains standing for 4 seconds. Fig.~\ref{fig:disturbance_stats} shows the results of this experiment. HLIP remained standing in only 30\% of trials, while our proposed approach achieved a 62\% success rate.

We randomized the timing of the push disturbance uniformly across the gait cycle. This blurs the boundary between falls and successes: the same disturbance might be more or less difficult depending when it is applied. 

\begin{figure} 
    \centering
    \includegraphics[width=\linewidth]{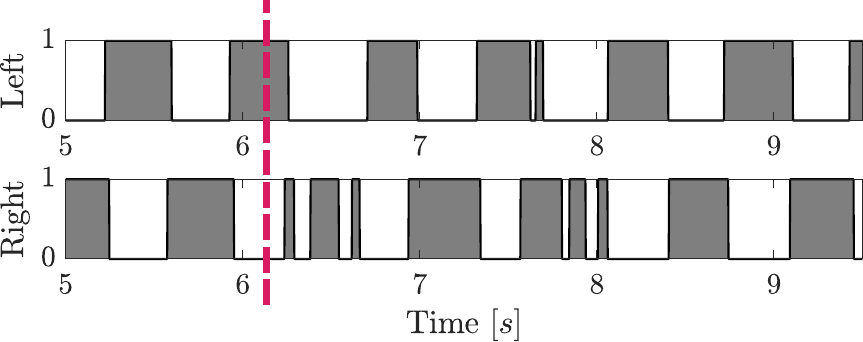}
    \caption{CI-MPC modulates the HLIP nominal contact sequence after being pushed. The dashed line indicates the time at which the disturbance is applied.}
    \label{fig:disturbance_contact}
    \vspace{-14pt} 
\end{figure}

Fig.~\ref{fig:disturbance_contact} shows the contact schedule that arises after a push disturbance. While the robot is tracking a forward velocity, a $0.2$ second disturbance of $\boldsymbol{f} = [50, \; -50, \; 0]^\T$ $\rm N$ is applied at the base frame of the robot. The proposed framework allows the robot to deviate from a nominal orbit and come back to it after recovery.

\begin{table}
\centering
    \begin{tabular}{c c c | c c}
        \begin{tabular}{@{}c@{}}Estimation\\Noise $\sigma_x$\end{tabular} & 
        \begin{tabular}{@{}c@{}}Link Mass\\Error $\sigma_m$\end{tabular} & 
        \begin{tabular}{@{}c@{}}Unmodeled\\Rough Terrain\end{tabular} & 
        HLIP only & 
        Proposed \\ 
        \hline
        0 & 0 & Yes & 1 / 10 & \textbf{7 / 10} \\
        0 & 0.2 & No & \textbf{10 / 10} & 9 / 10 \\
        0.05 & 0 & No & 0 / 10 & \textbf{8 / 10} \\
        0.05 & 0.2 & No & 0 / 10 & \textbf{3 / 10} \\
        0.05 & 0.2 & Yes & 0 / 10 & \textbf{1 / 10} \\
        0.01 & 0.1 & Yes & 0 / 10 & \textbf{7 / 10} \\
        \hline
    \end{tabular}
    \caption{Success rates, walking with model and estimation error.}
    \label{tab:uncertainty}
\end{table}

\subsection{Model and State Uncertainty on Rough Terrain}
Finally, while we have not yet validated HLIP-guided CI-MPC on hardware, the preliminary tests shown in Tab.~\ref{tab:uncertainty} indicate some level of robustness to state estimation and modeling error. In these experiments, we add Gaussian noise to the
state estimate,
\begin{equation}
    \hat{\bm x} \gets \hat{\bm x} + \bs \epsilon_x, \quad \bs \epsilon_x \sim \mathcal{N}(\bm 0, \sigma_x^2 \bm I),
\end{equation}
randomly alter the mass of each link,
\begin{equation}
    m_i \gets \max(0, m_i \cdot \epsilon_{m,i}), \quad \epsilon_{m,i} \sim \mathcal{N}(1, \sigma_m^2),
\end{equation}
and add randomly generated spheres (as in Fig.~\ref{fig:scenarios:terrain}) as unmodeled rough terrain. The robot is commanded to walk forward at 0.5 m/s, and a trial is counted as successful if the robot remains standing after 10 seconds. The HLIP-only controller, which has been validated on hardware \cite{ghansah2024dynamic}, is considerably less robust in all but one of these trials.


\section{Limitations} \label{sec:limitations}

While the above results demonstrate an important step toward flexible and robust humanoid locomotion, our proposed technique is not a panacea. 
Both HLIP and CI-MPC require parameter tuning, and their combination only increases the complexity of this process. While we used only one set of parameters for all the experiments, we did find that some parameters induced sharp tradeoffs. For example, a lower weight on base orientation tracking gave more natural-looking gaits, but reduced push recovery performance.

Our CI-MPC implementation uses significantly simplified collision geometries. This enables fast solve times, but precludes behaviors that involve contact away from the hands and the feet. As a result, the robot is not able to automatically recover from a fall. Furthermore, our CI-MPC solver's performance is reliant on smooth collision geometries, as sharp corners introduce problematic discontinuous gradients. 
Similarly, self-collisions present a major failure mode in the current implementation. Adding self-collision constraints either in the optimization problem \cite{grandia2021multi} or with a high order control barrier function \cite{khazoom2024tailoring, ames2019control, singletary2021safety} presents an obvious next step for improving reliability.

Finally, there are instances in which HLIP's suggested contact sequence guides the robot in an unhelpful direction. For example, if the robot is standing and pushed to the left, HLIP might suggest lifting the right leg, depending on the timing of the gait cycle. This could be mitigated with a richer reduced-order model, but illustrates a trade-off inherent to guiding whole-body behaviors with a reduced-order model.


\section{Conclusion}\label{sec:conclusion}
In this work, we proposed a method that leverages HLIP to guide CIMPC and allows the robot to make and break contact as needed. This framework enables diverse behavior for whole-body humanoid control like robust walking, disturbance recovery, and multi-contact stabilization. We have shown the effectiveness of this framework by testing it in simulation under disturbances, model and state uncertainty, and over adverse terrain. Future work will focus on realizing this framework on hardware as well as addressing the limitations delineated in Section \ref{sec:limitations}.

\balance
\clearpage

\bibliographystyle{ieeetr}
\bibliography{References/refs-zotero, References/refs-other}

\end{document}